\documentclass[letterpaper, 10 pt, conference]{ieeeconf}  

\IEEEoverridecommandlockouts                              

\usepackage{graphicx}
\usepackage{float}
\usepackage{multicol}
\usepackage{algorithmic}
\usepackage{algorithm}
\usepackage{multicol}

\title{\LARGE \bf
Backward Curriculum Reinforcement Learning
}

\author{KyungMin Ko$^{1}$
\thanks{*This work is supported by Georgia Tech’s summer undergraduate in engineering program (SURE), which is part of the National Science Foundation (NSF). Special thanks to professor Maguluri and graduate mentor Sajad Khodadadian for advising during the project.}
\thanks{$^{1}$Kyung Min Ko is with the department of electrical engineering, Purdue University, Indiana, USA,
        {\tt\small ko120@purdue.edu}}%
\thanks{In the proceedings of the 32nd IEEE International Conference on Robot and Human Interactive Communication (IEEE RO-MAN 2023)}
}

\begin{document}

\maketitle
\thispagestyle{empty}
\pagestyle{empty}


\begin{abstract}

Current reinforcement learning algorithms train an agent using forward-generated trajectories, which provide little guidance so that the agent can explore as much as possible. While realizing the value of reinforcement learning results from sufficient exploration, this approach leads to a trade-off in losing sample efficiency, an essential factor impacting algorithm performance. Previous tasks use reward-shaping techniques and network structure modification to increase sample efficiency. However, these methods require many steps to implement. In this work, we propose novel backward curriculum reinforcement learning that begins training the agent using the backward trajectory of the episode instead of the original forward trajectory. This approach provides the agent with a strong reward signal, enabling more sample-efficient learning. Moreover, our method only requires a minor change in the algorithm of reversing the order of the trajectory before agent training, allowing a straightforward application to any state-of-the-art algorithm.

\end{abstract}

 \begin{figure*}[!ht]
      \centering
      \includegraphics[width =\textwidth]{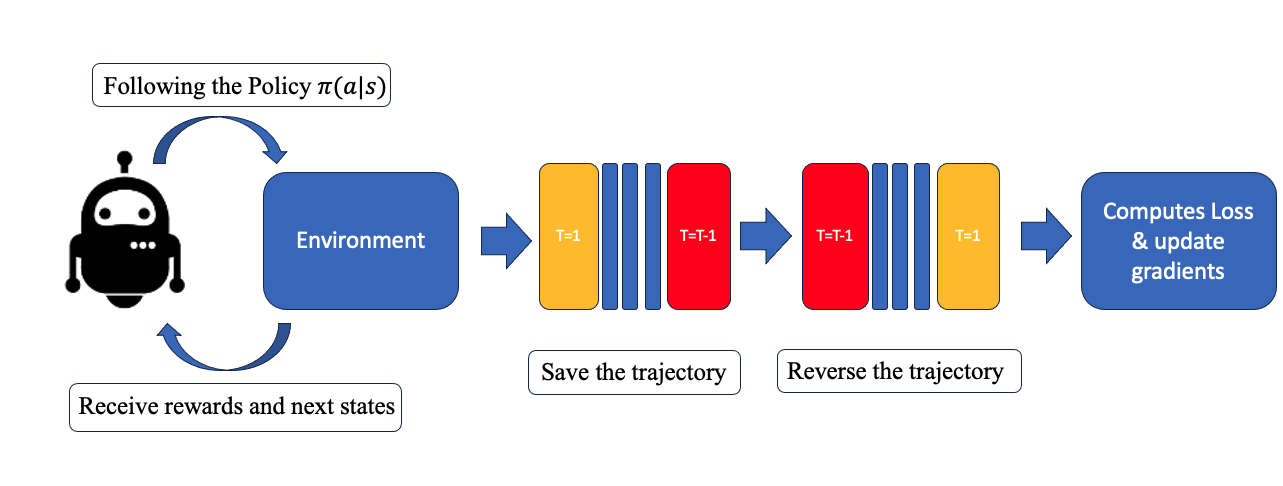}
      \caption{Workflow of Backward Curriculum Algorithm.}
      \label{figurelabel}
   \end{figure*}
\section{INTRODUCTION}

Recent developments in GPU-based computation have enabled tremendous advances within the field of deep learning. One result has extended the implementation of reinforcement learning (RL) to various fields. As a powerful technique that trains an agent to maximize some reward gain to solve a computational problem efficiently, RL is applied in many applications, such as robotic manipulation tasks \cite{multiagentsystem2}, playing Atari games \cite{atarigame}, controlling multiagent systems \cite{roboticmanupulation1}, and learning AlphaGo \cite{alphago}. However, the natural reward function of an RL algorithm is usually sparse because a reward is given to the agent only when it completes a given task. Moreover, conventional RL problems are formulated with the agent being blind to the task goal. While this configuration helps the agent discover an optimal policy without human guidance, significant computational power and a large sample size are often required for training. 

In our work, we introduce an approach of novel backward curriculum reinforcement learning that uses agent trajectories in reverse order for training. This enables the agent to begin training by recognizing the task goal. As a result, the agent trains in a sample efficient way by utilizing strong reward signals during backward curriculum learning. Our method requires no modification of the neural network structure and no other previous knowledge to train the agent. Moreover, it can be simply applied to any state-of-the-art algorithm, such as PPO \cite{ppo3}, A3C \cite{a3c4}, and SAC \cite{sac5}, with two straightforward steps. The trajectory of the agent is collected first, as usual, then the order of training the agent is reversed as shown on figure 1. 
Our backward curriculum learning leverages the concept of curriculum learning, which manually changes the order of the training process to train more efficiently. We empirically test this method using the REINFORCE \cite{reinforce7} and REINFORCE with baseline \cite{reinforcebaseline} algorithms on the CartPole-v1 and Lunar Lander-v2 environments from the OpenAI gym \cite{openaigym10} framework, which both use discrete action spaces. Moreover, we further investigate the effects of the return normalization technique \cite{returnnorm9}, variation of the learning rate, and structure of neural network on the performance of our method.

\section{Related Works}

Some published approaches share similar features with our backward curriculum learning method. First, imitation learning \cite{imitationlearning11} requires the demonstration of experts who explain to the agent how to reach the goal task, which can use samples efficiently to train agents. However, imitation learning limits agents from taking advantage of exploration. Moreover, significant effort is required because experts who can demonstrate completing the task and record the process to translate for agent training must be employed. On the other hand, our method does not necessitate additional effort, as we only need to reverse the trajectory for training. Compared to this approach, we use our backward trajectory as the expert seen in imitation learning. 

Curriculum learning modifies the schedule of the learning process and has been applied to various machine learning tasks. The curriculum learning approach first trains the agent on easier examples, then continuously increases the difficulty level for solving the problem \cite{curriculum6}. The application of curriculum learning for helping children to learn chess called "Quick Chess" helped the children to efficiently learn chess by giving a sequence of progressively more difficult games \cite{curriculum6}. The concept of curriculum learning was first applied to the artificial intelligent domain back in the 1990s \cite{curriculum6}. Where the first application is known as grammar learning \cite{grammer}. In backward curriculum learning, we train the agent from the end to the beginning of the episode, which follows the concept of curriculum learning because the agent can complete the task from near the goal state more easily than the start state. Previously, curriculum learning was applied to pre-specified tasks, such as shooting a ball into a goal \cite{curriculumlearningshootingball}. Barnes proposed similar work to solve complex robotic problems \cite{difficultrobot}. However, this method requires partitioning the entire task space, which limits its application to various problems. On the other hand, our method is applicable to a range of machine learning tasks because previous knowledge is not required to train the agent. 

Moreover, we even further increased the performance of our algorithm by adopting the return normalization method. In the domain of reinforcement learning, the modern architecture of deep neural networks mostly relies on feed forward neural networks. Compared to large capacity models from computer vision and natural language processing domain, reinforcement learning instead adopts algorithmic development such as novel loss function to achieve state of art performance \cite{returnnormhist}. Researchers also have proven that increasing the model capacity in reinforcement learning will harm the performance \cite{largemodelRL}. In reinforcement learning, the cumulative sum of rewards often ends up with a high magnitude value, so it will increase the variance during the training process \cite{returnnorm9}. In reinforcement learning, the role of variance is significant since high variability will lead the agent to an unwanted local minimum point \cite{variance}. Therefore, applying return normalization technique will generate less variance environment for training.

\section{Preliminaries}

We consider a finite discrete time horizon Markov decision process (MDP) as a tuple \(M=(S,A,P,r,\gamma,\rho_{0},T\)), with \(S\) defined as a state set, \(A\) as an action set, and \(P: S\times A\times S \rightarrow R\) as a transition probability. Also, \(r\) is a bounded reward function, \(\gamma\) is a discounted reward factor, \(\rho_{0}\) is the initial state distribution, and \(T\) is the trajectory of the moving agent. At each time step in MDP, the agent takes an action, receives a reward, and moves to the next state using the transition probability, \(P\). The goal of RL in this setup is to find the optimal policy, \(\pi_{\theta}(a_{t}|s_{t})\), that maximizes the reward gain. 
In our backward curriculum learning approach, we reverse the order of the agent trajectory, \(T\), labeled as \(T_{b}\), and use this to train the agent from the goal state, \(s_{g}\), to the initial state, \(s_{0}\). Because the agent begins training near \(s_{g}\),  the agent receives a strong reward signal, providing meaningful guidance toward the goal.

\section{Backward Curriculum Learning}

\subsection{Backward Curriculum Learning in REINFORCE}

The REINFORCE algorithm is one basic policy gradient algorithm that provides insights for building state-of-art policy gradient algorithms. We first applied backward curriculum learning to this algorithm because it is a basic policy gradient algorithm and is intuitive and easy to implement. The REINFORCE algorithm includes a simple structure that uses cumulative discounted returns and the log probability of choosing an action to compute gradients. However, the original REINFORCE algorithm features a sparse reward function that lowers its performance on complicated tasks. Backward curriculum learning can solve this problem by enabling the agent to begin updating its gradient from the end of the episode. Then, by applying our backward curriculum learning, the agent recognizes the goal at the beginning of the training process, which replaces the original sparse reward function with a strong reward signal. Detailed in Algorithm 2, this approach first collects the sample trajectory using the policy network, then flips the order of the episodes to compute the loss function in reverse order which is different from Algorithm 1 that uses trajectory collected by forward sequence.

\begin{algorithm}[ht]
   \caption{REINFORCE} 
\begin{algorithmic}
   \STATE Collect the sample trajectories following $\pi_{\theta}(a_{t}|s_{t})$
   \STATE $T=eqepisode \;length$
   \FOR{$t=1$ {\bfseries to} $t=T-1$}
   \STATE Compute Return $G$
   \STATE $\theta \leftarrow \theta_{old} + G \nabla(\log\pi_{\theta}(a_{t}|s_{t}))$ 
   \STATE Update $Optimizer$
   \ENDFOR

\end{algorithmic}
\end{algorithm}

\begin{algorithm}[ht]
   \caption{Backward Curriculum REINFORCE}
\begin{algorithmic}
   \STATE Collect the sample trajectories following $\pi_{\theta}(a_{t}|s_{t})$
   \STATE $T=eqepisode \;length$
   \FOR{$t=T-1$ {\bfseries to} $t=1$}
   \STATE Compute Return $G$
   \STATE $\theta \leftarrow \theta_{old} + G \nabla(\log\pi_{\theta}(a_{t}|s_{t}))$ 
   \STATE Update $Optimizer$
   \ENDFOR

\end{algorithmic}
\end{algorithm}

\subsection{Backward Curriculum Learning in REINFORCE with Baseline}
A key challenge associated with the REINFORCE algorithm is dealing with a high variance that may cause a divergence of policy network parameters. The typical solution for reducing the variance is subtracting the baseline \cite{reinforcebaseline}. In the REINFORCE algorithm, a value function is an appropriate baseline to subtract from the returned G. The agent has the policy network and value network, and the loss function is $\theta \leftarrow \theta_{old} + (G-V) \nabla(\log\pi_{\theta}(a_{t}|s_{t}))$ as shown on Algorithm 3. This baseline decreases the variance by reducing the step size of the gradient. In our backward REINFORCE with baseline algorithm, we first collect the trajectories using our policy, then flip the order of the episode to compute the loss function, as shown in Algorithm 4.

\begin{algorithm}[ht]
   \caption{REINFORCE with Baseline}
   \label{algorithm3}
\begin{algorithmic}
   \STATE Collect the sample trajectories following $\pi_{\theta}(a_{t}|s_{t})$
   \STATE $T=eqepisode \;length$
   \FOR{$t=1$ {\bfseries to} $t=T-1$}
   \STATE Compute Return $G$
   \STATE $\theta \leftarrow \theta_{old} + (G-V) \nabla(\log\pi_{\theta}(a_{t}|s_{t}))$ 
   \STATE Update $Optimizer$
   \ENDFOR

\end{algorithmic}
\end{algorithm}
\begin{algorithm}[ht]
   \caption{Backward Curriculum REINFORCE with Baseline}
   \label{algorithm4}
\begin{algorithmic}
   \STATE Collect the sample trajectories following $\pi_{\theta}(a_{t}|s_{t})$
   \STATE $T=eqepisode \;length$
   \FOR{$t=T-1$ {\bfseries to} $t=1$}
   \STATE Compute Return $G$
   \STATE $\theta \leftarrow \theta_{old} + (G-V) \nabla(\log\pi_{\theta}(a_{t}|s_{t}))$ 
   \STATE Update $Optimizer$
   \ENDFOR

\end{algorithmic}
\end{algorithm}

\section{Experimental Result}
We incorporated the RL testing environment from OpenAI Gym \cite{openaigym10} to test the performance of our backward curriculum learning algorithm on the REINFORCE and REINFORCE with baseline algorithms. For these experiments, we evaluated with the Cart Pole-v1 and Lunar Lander-v2 environments. The default structure of the network is a multi-layer perceptron network with two layers and 128 neurons as a baseline. Moreover, we analyzed the effects of return normalization and modifying the network structure. 
 \begin{figure*}[!ht]
      \centering
      \includegraphics[width =\textwidth]{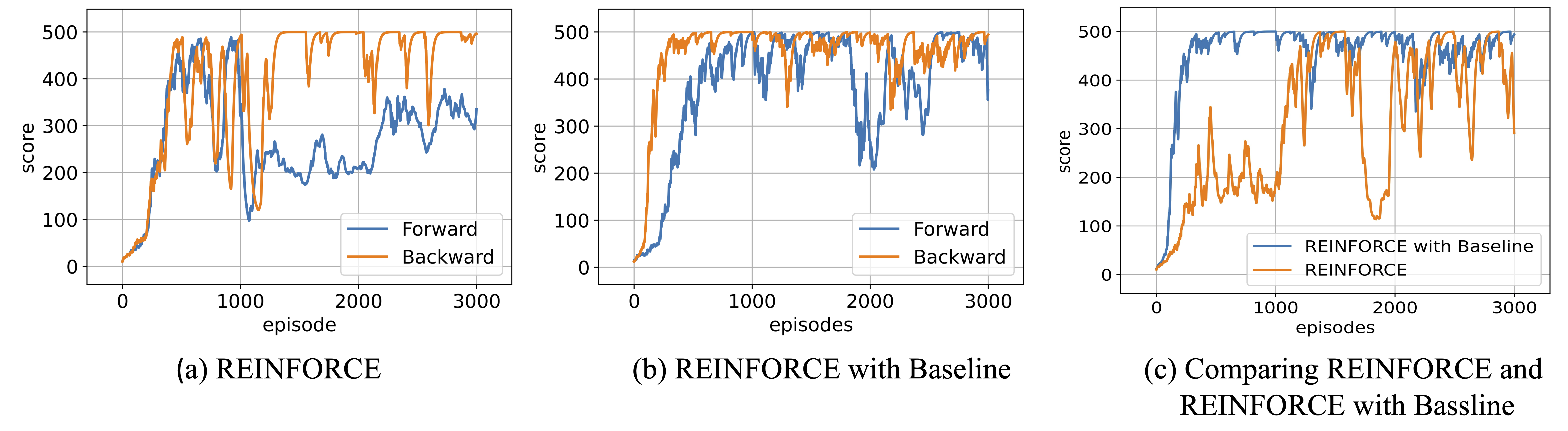}
      \caption{Experimental results with the Cart Pole environment.}
      \label{figurelabel}
   \end{figure*}

 \begin{figure*}[!ht]
      \centering
      \includegraphics[width = \textwidth]{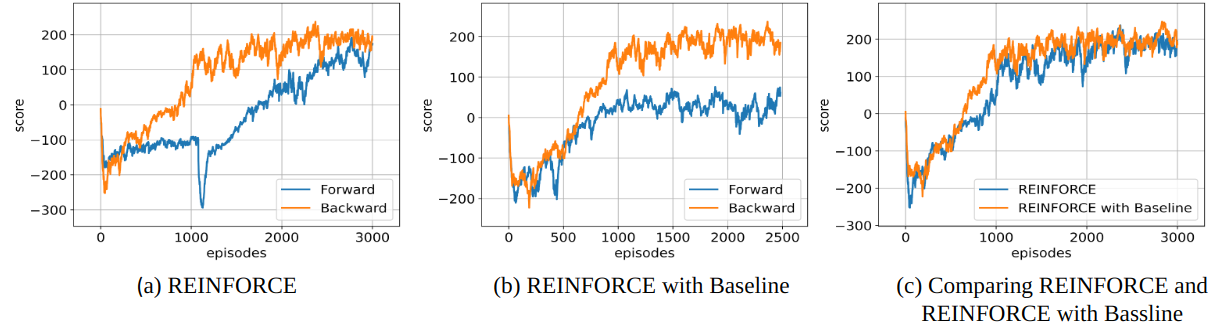}
      \caption{Experimental results with the Lunar Lander environment.}
      \label{figurelabel}
   \end{figure*}
\subsection{Cart Pole Environment}
We first implemented our backward curriculum method to the REINFORCE algorithm with the goal of the Cart Pole-v1 algorithm to balance a pendulum on a cart by applying a discrete action domain with the force of +1 or -1 to the cart. The environment is solved when the agent reaches a score of 500 without the pole falling from the cart.

A comparison of the backward curriculum algorithm and the original REINFORCE algorithm is shown in Figure 2(a). No apparent difference occurs with the first 1,000 episodes. However, after this point, the backward curriculum algorithm begins to outperform the original algorithm and solves the environment within 3,000 episodes. 

Next, we experimented using the REINFORCE with a baseline algorithm to analyze the effect of the backward curriculum learning algorithm. In Figure 2(b), the backward curriculum algorithm solves the environment within 250 episodes. On the other hand, the original REINFORCE with the baseline algorithm requires more episodes to solve and has a high variance that disturbs the agent from remaining in the goal state. We further observe that adding a baseline to the REINFORCE algorithm effectively solved a high variance problem while improving the performance of the algorithm. We see that solving the environment using the REINFORCE algorithm takes about 1,500 episodes, and adding the baseline results requires less than 500 episodes, as shown in Figure 2(c).

\subsection{Lunar Lander Environment}

We next explored the Lunar Lander-v2 environment from the OpenAI gym with the agent goal of safely landing within a specified region without crashing. The action space consists of the four actions of resting, firing left, firing right, and firing the main engines. The environment is solved when the agent reaches a score of 200 without crashing. Here, we compare the performance of the backward curriculum and original algorithms and experiment with the effect of return normalization. 

Figure 3(a) shows that our backward curriculum REINFORCE algorithm reaches the goal state with an average score of 200 within 2,000 episodes, while the REINFORCE algorithm solves the environment within 3,000 episodes. This result suggests that enabling the agent to know the environment goal at the beginning of the training process significantly helps it to reach the goal using fewer samples. Also, in Figure 3(b), the REINFORCE with baseline using the backward curriculum learning algorithm completes the task faster in 1,000 episodes using the reverse method, while the original algorithm requires more than 2,500 episodes. We compare the performance of the backward curriculum REINFORCE and backward curriculum REINFORCE with the baseline in Figure 3(c), which shows both methods attain the goal score. However, the backward curriculum learning on REINFORCE with baseline features less variance and uses fewer samples.

\subsection{Return Normalization}
\label{Return Normalization}
Backward curriculum learrning on the REINFORCE algorithm gradually optimizes the sample efficiency but still includes the problem of high variance. Considering the original REINFORCE loss function of $\theta \leftarrow \theta_{old} + G \nabla(\log\pi_{\theta}(a_{t}|s_{t}))$, the gradient step size depends on $G$ and log $\log\pi_{\theta}(a_{t}|s_{t})$. The log of the action probability is usually acceptable, but the return has a high magnitude that will dynamically increase the step size of the gradient. The choice of step size is crucial in RL because a small step size slows the convergence rate, and a large step size may cause oscillations or divergence of policy networks due to overshooting \cite{overshooting_problem}.

Therefore, we apply return normalization to our method to stabilize the observed high variance by scaling the mean of the return to zero and the variance to one, which reduces the magnitude of the return and avoids overshooting. We implement this by subtracting the average of the return and dividing by the standard deviation of the return, following Equation 1 \cite{returnnorm9}. Applying return normalization results in a stable backward curriculum REINFORCE algorithm with increased performance.

We implement this return normalization approach on the original algorithm and our backward curriculum algorithm simulated with the Lunar Lander environment. As observed in Figure 4(a), before applying return normalization, the learning curve features a high variance, and some learning rates result in divergence. Then, including return normalization decreases the variance significantly, and an agent with a learning rate of $e^{-4}$ solves the environment within 3,000 episodes, as shown in Figure 4(b). This outcome demonstrates that return normalization optimizes the performance and minimizes the variance, as seen in Figure 4(c), comparing with and without return normalization using the best-performing learning rate of $e^{-4}$.

\begin{equation}
Return\:Norm = \frac{return - average\:return}{standard\:deviation \: of \: return }
\end{equation}

 \begin{figure*}[ht]
      \centering
      \includegraphics[width = \textwidth]{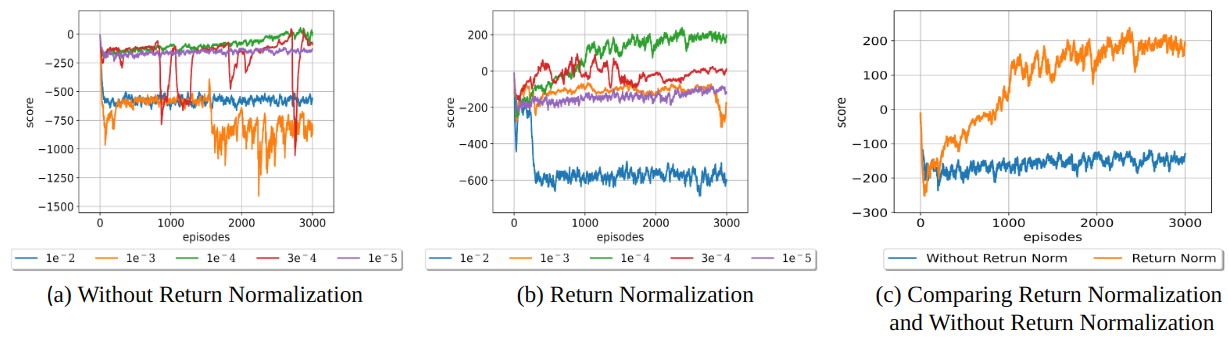}
      \caption{Effect of return normalization on backward curriculum learning with the Lunar Lander environment.}
      \label{figurelabel}
   \end{figure*}

\begin{figure*}[ht]
      \centering
      \includegraphics[width = 0.6\textwidth]{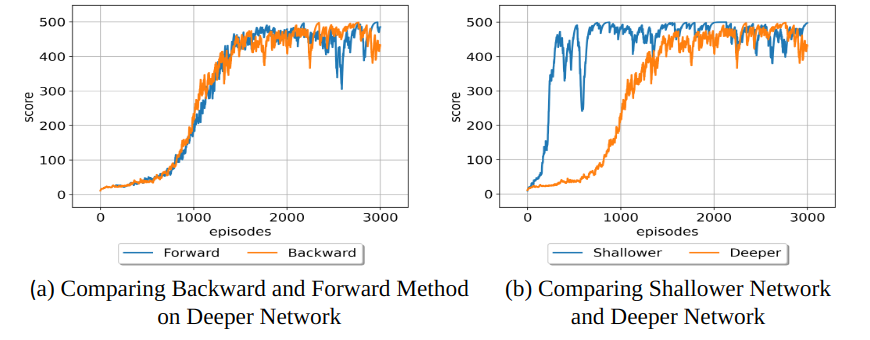}
      \caption{Effect of the deep neural network depth on backward curriculum learning with the Cart Pole environment.}
      \label{figurelabel}
   \end{figure*}
\subsection{Comparison between deep and shallow networks}
The depth of the neural network is an important factor that affects the performance of a deep learning algorithm. A deeper neural network typically requires more computational power because increasing the number of layers yields more parameters with which to compute the gradient. Instead, a shallow network offers the benefit of quicker computation because of its fewer parameters from fewer layers. While deep and shallow networks offer distinct benefits, we observe that the shallow network performs better for simple environments, such as Cart Pole. In this scenario, the agent can solve the problem through simple heuristics without wasting computation efforts \cite{deep_shallow}. If we use deeper networks to solve this simple environment, then that agent may “overthink,” resulting in computational waste and potential misclassification \cite{deep_shallow}.
 
Our experiments use shallow and deep networks to compare algorithm performance with the Cart Pole environment. The baseline network structure contains two layers with 128 neurons, and the deeper network for comparison includes three layers with 256 neurons. We applied backward curriculum REINFORCE with baseline and the original REINFORCE with baseline method to observe differences with the deeper network. 

The backward curriculum method does not improve the performance, as we otherwise expected, which is shown in Figure 5(a). This result suggests that a deeper network already forms an accurate approximation, so enabling the agent to start the training process with an awareness of the task goal offers little impact on performance. Comparing the performance between the shallow and deep networks using the Cart Pole environment, the shallow network agent completes the task in fewer than 500 episodes, as shown in Figure 5(b). However, the deeper network agent takes over 1,500 episodes to reach the goal state. Therefore, we suggest that backward curriculum learning is most effective in a shallow network configuration to solve simple tasks by maximizing sample efficiency. 

\section{Conclusion}
We proposed a novel backward curriculum reinforcement learning to address the natural sparse reward function problem. Our method reverses the order of the episode before beginning the training process, enabling the agent to be initialized with a recognition of the task goal. The natural sparse reward function is replaced with a strong reward signal, which optimizes the sample efficiency. Unlike previously proposed methods, backward curriculum learning does not require many steps to modify the code structure, so we can directly apply our method to state-of-the-art algorithms. We selected the REINFORCE and REINFORCE with a baseline algorithms to test our method because these represent the state of the art. We empirically demonstrated that backward curriculum learning uses fewer samples to solve a given task. Also, we tested the effect of return normalization and network depth with our backward curriculum learning algorithm approach and observed that the backward curriculum learning is appropriate for solving simple tasks, like Cart Pole, using a shallow network. In future work, we will apply our backward curriculum learning on other state-of-the-art algorithms to test on various environments.

\bibliographystyle{IEEEtran}
\bibliography{IEEEabrv,main}
\end{document}